\documentclass{article}

 \usepackage[preprint]{neurips_2025}

\usepackage[utf8]{inputenc} % allow utf-8 input
\usepackage[T1]{fontenc}    % use 8-bit T1 fonts
\usepackage{hyperref}       % hyperlinks
\usepackage{url}            % simple URL typesetting
\usepackage{booktabs}       % professional-quality tables
\usepackage{amsfonts}       % blackboard math symbols
\usepackage{nicefrac}       % compact symbols for 1/2, etc.
\usepackage{microtype}      % microtypography
\usepackage{xcolor}         % colors
\usepackage{graphicx}
\usepackage{subcaption}
\usepackage{array}
\usepackage{amsmath}
\usepackage{multirow}
\title{Pack and Force Your Memory: Long-form and Consistent Video Generation}

\author{%
  Xiaofei Wu$^{1,2,}$\thanks{Work is done during internship at Tencent Hunyuan.} \\
  \And
  Guozhen Zhang$^{3,2,*}$ \\
  \And
  Zhiyong Xu$^{2}$ \\
  \And
  Yuan Zhou$^{2,}$\thanks{Project leader.} \\
  \And
  Qinglin Lu$^{2}$ \\
  \And
  Xuming He$^{1,}$\thanks{Corresponding author.} \\
  \AND
  $^1$ ShanghaiTech University   $^2$ Tencent Hunyuan   $^3$ Nanjing University 
}

\begin{document}

\maketitle

\begin{abstract}
Long-form video generation presents a dual challenge: models must capture long-range dependencies while preventing the error accumulation inherent in autoregressive decoding. To address these challenges, we make two contributions. First, for dynamic context modeling, we propose MemoryPack, a learnable context-retrieval mechanism that leverages both textual and image information as global guidance to jointly model short- and long-term dependencies, achieving minute-level temporal consistency. This design scales gracefully with video length, preserves computational efficiency, and maintains linear complexity. Second, to mitigate error accumulation, we introduce Direct Forcing, an efficient single-step approximating strategy that improves training–inference alignment and thereby curtails error propagation during inference. Together, MemoryPack and Direct Forcing substantially enhance the context consistency and reliability  of long-form video generation, advancing the practical usability of autoregressive video models. Project website: \url{https://wuxiaofei01.github.io/PFVG/}.
\end{abstract}

\section{Introduction}

Video generation has emerged as a central problem in generative modeling, enabling applications in content creation~\citep{hu2025HunyuanVideo-Avatar}, embodied intelligence~\citep{gr1,liu2024realdex,gr2,Wu2024FastGraspEG}, and interactive gaming~\citep{team2025yan,yu2025context,Li2025HunyuanGameCraftHI,yu2025gamefactory}. Recent  Diffusion Transformer (DiT) models~\citep{kong2024hunyuanvideo,wan2.2,seedance,waver} demonstrate strong capabilities in capturing complex spatiotemporal dependencies and character interactions within fixed-length sequences, producing realistic video clips. However, long-form video generation remains challenging: the substantially larger token count of videos renders end-to-end modeling with quadratic-complexity DiT architectures computationally prohibitive, and the lack of effective long-term modeling leads to increasingly severe drift as video length grows. These factors pose significant challenges for generating minute-scale or longer videos while maintaining temporal coherence and computational efficiency.

Existing approaches~\citep{ai2025MAGI1AV,Zhang2025PackingIF} typically enhance context consistency by retaining only the most recent frames or applying fixed compression strategies to select key frames. However, due to the limited window size and aggressive token compression, these rigid mechanisms rely primarily on local visual information and fail to capture global dependencies. In long-form video generation, the absence of global context inevitably degrades temporal coherence.

We address this challenge by reformulating long video generation as a long-short term information retrieval problem, where the model must effectively retrieve both persistent long-term context and dynamic short-term cues to guide frame synthesis. Specifically, we introduce MemoryPack, a linear-complexity dynamic memory mechanism that leverages textual and image information as global guidance. MemoryPack retrieves long-term video context that is semantically aligned with the overall narrative to reinforce temporal coherence, while simultaneously exploiting adjacent frames as short-term cues to enhance motion and pose fidelity. In contrast to methods based on fixed compression or frame selection, MemoryPack enables flexible associations between historical information and future frame generation.

Another central challenge in long-form video generation is error accumulation caused by the training--inference mismatch: during training, models condition on ground-truth frames, whereas at inference they rely on self-predictions, causing errors to compound over long horizons. Although \cite{Zhang2025PackingIF} attempts to mitigate drift by generating frames in reverse order, the reliance on tail frames may reduce motion dynamics in long videos, limiting its effectiveness. \cite{selfforce} addresses this by conditioning directly on generated outputs, but to improve efficiency it requires distribution distillation, which introduces extra computation and, due to the inherent limitations of distillation, the generation quality is degraded, and consequently, incorporating previously generated results into the training process introduces additional noise.

% degrades generation quality and injects additional noise during training.

To address this issue, we introduce Direct Forcing, an efficient strategy that aligns training with the model’s inference in a single step. Building on rectified flow~\citep{rectifiedflow}, we perform one-step backward ODE computation in the predicted vector field to approximate inference outputs. This method incurs no additional overhead, requires no distillation, preserves train–inference consistency, and mitigates error accumulation.

Our method achieves state-of-the-art performance on VBench~\citep{huang2023vbench} across key metrics, including Motion Smoothness, Background Consistency, and Subject Consistency, while further enhancing robustness against error accumulation. Experimental results demonstrate that MemoryPack and Direct Forcing effectively model long-term contextual information and achieve superior consistency performance.

In summary, our contributions are threefold:
\begin{itemize}
    \item \textbf{MemoryPack}: a dynamic memory mechanism that leverages text and image as global guidance to retrieve historical context, while exploiting adjacent frames as short-term cues. This design enables efficient modeling of minute-level temporal consistency without relying on rigid compression.
    \item \textbf{Direct Forcing}: a single-step approximating strategy that aligns training with inference efficiently, eliminating distillation and mitigating error accumulation in long-horizon generation.
    \item \textbf{State-of-the-Art Performance}: extensive evaluations demonstrate that our approach achieves state-of-the-art results in long-term consistency metrics.
\end{itemize}

\section{RELATED WORKS}
% 长视频生成仍然是一个悬而未决的问题。由于大多数现有方法仍然建立在DiT框架之上，它们受到$\mathcal{O}（L^2）$计算复杂性的影响，其中$L$表示上下文标记的数量，对于长序列来说，它变得过于庞大。为了解决这个问题，已经提出了几种策略。一些工作用线性的注意力取代了原来的机制，以提高效率。TTT通过测试时间训练来减轻训练成本，而Self-Force通过使用KV-cache进行滚动更新来加速训练。此外，一种无需训练的方法SparseVideoGen（SVG）利用了3D完全注意力的固有稀疏性，以进一步提高推理效率。
\subsection{Efficient Accumulate for Long Video Generation}
Long video generation remains an open problem. Since most existing approaches are still built upon the DiT framework, they suffer from the $\mathcal{O}(L^2)$ computational complexity, where $L$ denotes the number of context tokens, which becomes prohibitively large for long sequences. To address this issue, several strategies have been proposed. Some works replace the original mechanism with linear attention~\citep{l-att1,l-att2,l-att3,l-att4,peng2023rwkv,gu2023mamba} to improve efficiency. \cite{ttt} alleviates training costs through test-time training, while \cite{selfforce} accelerates training by employing a KV-cache for rolling updates. In addition, a training-free approach, \cite{xi2025sparse} (SVG), exploits the inherent sparsity of 3D full attention to further enhance inference efficiency.

% 当前的视频生成研究主要集中在秒级持续时间上。最近，几项研究开始探索如何扩展秒级模型以生成分钟长的视频。例如，方法A和B采用DiT作为骨干网络，并以先前生成的帧为条件迭代预测未来帧。然而，这些方法在建模长期依赖关系方面面临挑战。具体而言，A和B将模型限制为只关注最近的帧，这提高了计算效率，但不可避免地导致长期信息的丢失。相比之下，方法C引入了MOC方案，该方案通过路由选择上下文信息。虽然这提高了利用上下文的灵活性，但它依赖于手动设计的选择策略，从而限制了模型自主确定相关信息的能力。

\subsection{Framework for Video Generation}
Current video generation research~\citep{zhang2025motion,seedance,waver,wan2.2,kong2024hunyuanvideo,yang2024cogvideox,opensora,opensora2} has primarily focused on short clips at the second level. Recently, several studies have explored extending such models to generate minute-long videos. For example, \cite{Zhang2025PackingIF} and \cite{ai2025MAGI1AV} adopt DiT as the backbone network and iteratively predict future frames conditioned on previously generated ones. However, these approaches face challenges in modeling long-term dependencies. In particular, they restrict the model to attend only to the most recent frames, which improves computational efficiency but inevitably leads to the loss of long-range information. In contrast, \cite{moc} introduces the Mixture of Contexts (MoC), which selects contextual information through routing. While this design improves flexibility in leveraging context, it relies on manually defined selection rules, thereby limiting the model’s ability to autonomously determine relevant information.

\begin{figure}[t] % [t] 表示放在页面顶部
    \centering
    \includegraphics[width=\textwidth]{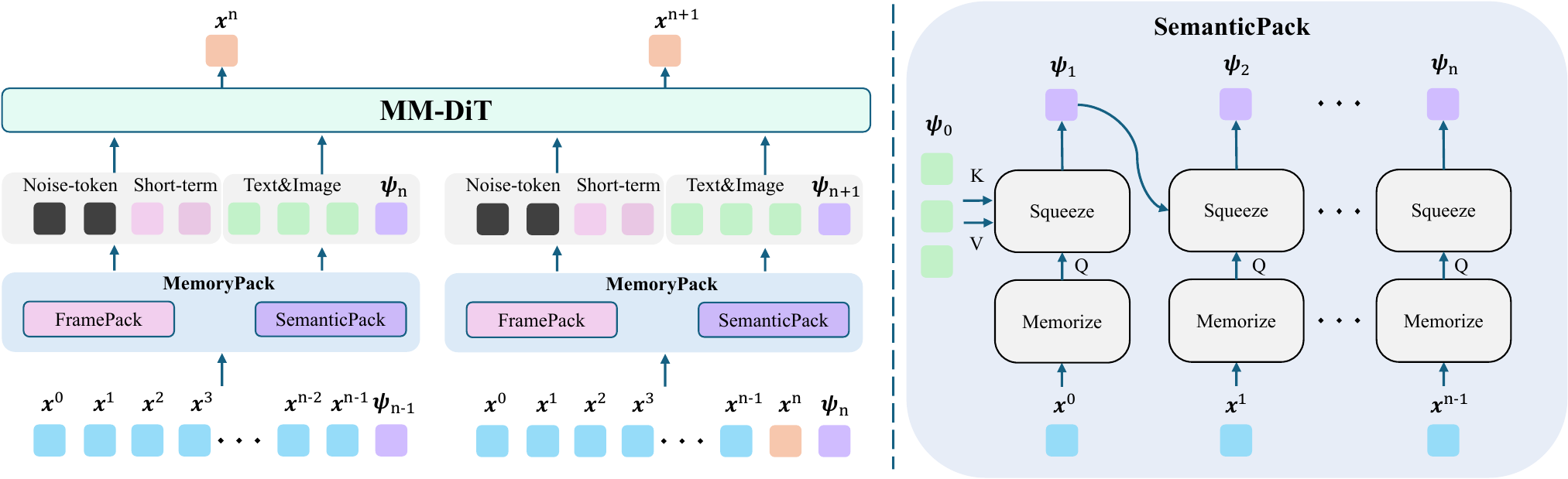}
    % 给一个文本和图片输入，我们首先会通过Memory-Pack和Frame—Pack进行层次化处理，获取得到长期和短期信息。在Memory-Pack中，将long term与文本信息进行concate作为语义信息，同时将short term与视觉信息进行concat作为视觉信息，最终再利用MM-Dit进行特征融合与预测。
    \caption{\textbf{Overview of our framework.} Given a text prompt, an input image, and history frames, the model autoregressively generates future frames. Prior to feeding data into MM-DiT, MemoryPack retrieves both long- and short-term context. In SemanticPack, visual features are extracted within local windows via self-attention, followed by cross-attention to align them with global textual and visual information to iteratively generate long-term dependencies $\psi_{n}$. This design achieves linear computational complexity and substantially improves the efficiency of long-form video generation.}
    \label{fig:framework}
\end{figure}
\section{Method}

Given $n$ historical segments ${\{\mathbf{x}^0, \dots, \mathbf{x}^{n-1}}\}$, along with a textual prompt $P$ and a conditional image $I$, our objective is to generate the subsequent segment $\mathbf{x}^n$. We formulate this as an autoregressive image-to-video generation task, enabling the synthesis of arbitrarily long video sequences from both textual and image inputs. Unless otherwise specified, all references to segments in this paper denote their latent-space representations. Our approach builds upon the Diffusion Transformer (DiT) architecture~\citep{dit,kong2024hunyuanvideo} to create an autoregressive model for future segments generation, with its overall architecture illustrated in Figure~\ref{fig:framework}. To enhance temporal consistency and mitigate error accumulation—inherent challenges in autoregressive generation—we introduce two key innovations: (i) MemoryPack, a hierarchical and efficient fusion module that leverages the text prompt and image as a global guide to model both long- and short-term temporal dependencies (Sec.~\ref{sec:MemoryPack}); and (ii) the Direct Forcing strategy, which aligns the training process with inference by employing single-step approximation. This approach mitigates the discrepancy between conditioning on ground-truth segments during training and on model-generated segments during inference (Sec.~\ref{sec:Direct Forcing}).

\subsection{MemoryPack}
\label{sec:MemoryPack}
Generating long-form videos requires balancing high-fidelity, smooth local motion with semantic coherence across the global narrative. Previous works~\citep{ai2025MAGI1AV,apt2}  typically condition video generation on a fixed sliding window of recent segments $\{\textbf{x}^{n-k}, \dots, \textbf{x}^{n-1}\}$ together with a text prompt $P$ and a conditional image $I$. While such methods excel at preserving local motion, they often fail to capture long-range dependencies, including object identities and scene layouts. Conversely, models focusing solely on distant context may lose track of fine-grained motion cues, leading to artifacts and temporal discontinuities.

To address these challenges, we introduce MemoryPack, a hierarchical module that jointly leverages complementary short-term and long-term contexts for video generation. It consists of two components: FramePack~\citep{Zhang2025PackingIF} and SemanticPack.

FramePack focuses on short-term context by capturing appearance and motion through a fixed compression scheme, thereby enforcing short-term consistency. However, its fixed window size and compression ratio constrain its ability to dynamically propagate information over long time horizons.

To maintain global temporal coherence, SemanticPack integrates visual features with textual and image guidance, unlike prior methods~\citep{moc} that rely solely on visual representations. This is achieved by iteratively updating a long-term memory representation $\boldsymbol{\psi}$  using contextual video segments $\{\textbf{x}^{0}, \dots, \textbf{x}^{n-1}\}$ , a text prompt $P$ , and a reference image $I$. The process consists of two structured operations: (1) Memorize, which applies self-attention within windows of historical segments to produce compact embeddings. This approach mitigates the prohibitive quadratic complexity of attending to long histories while retaining holistic window-level cues. (2) Squeeze, which then injects the textual and image guidance into this visual memory. Following prior work~\citep{wan2.2}, we implement this as a cross-attention layer where the output of Memorize serves as the query, and the representation $\boldsymbol{\psi}$ acts as the key and value. This alignment ensures the long-term memory remains globally aware and semantically grounded:
\begin{equation}
    \boldsymbol{\psi}_{n+1} = \text{Squeeze}\big(\boldsymbol{\psi}_{n}, \text{Memorize}(\textbf{x}^{n})\big).
\end{equation}
For initialization ($n=0$), we set $\boldsymbol{\psi_0}$ as the concatenation of the prompt feature and the reference image feature, providing a semantic prior that anchors the memory trajectory. Importantly, the computational complexity of SemanticPack is $\mathcal{O}(n)$, ensuring scalability by preventing costs from growing prohibitively with the number of historical frames. By integrating both short-(FramePack) and long-term(SemanticPack) context, MemoryPack retains information from distant frames while preserving temporal and motion consistency with high computational efficiency. Additional experiments on SemanticPack are provided in the appendix.

\paragraph{RoPE Consistency} In DiT-based autoregressive video generation, long videos are typically partitioned into multiple segments during training. However, this segmentation causes even adjacent segments from the same video to be modeled independently, leading to the loss of cross-segment positional information and resulting in flickering or temporal discontinuities. To address this issue, we treat the input image as a CLS-like token and incorporate RoPE~\citep{rope} to explicitly encode relative positions across segments. Specifically, during training, for each video clip, we assign the image the initial index of the entire video, thereby preserving coherence and enhancing global temporal consistency. Formally, RoPE satisfies
\begin{equation}
R_q(\boldsymbol{x}_q, m) R_k(\boldsymbol{x}_k, n)
= R_g(\boldsymbol{x}_q, \boldsymbol{x}_k, n-m), \quad
\Theta_k(\boldsymbol{x}_k, n) - \Theta_q(\boldsymbol{x}_q, m)
= \Theta_g(\boldsymbol{x}_q, \boldsymbol{x}_k, n-m),
\end{equation}
where $R_q, R_k, R_g$ denote the rotation matrices for query, key, and relative position, respectively; $\Theta_q, \Theta_k, \Theta_g$ denote the corresponding rotation angles. $\boldsymbol{x}_q$ and $\boldsymbol{x}_k$ are the query and key vectors, with $m$ and $n$ being their position indices.By assigning the image token an index of start, the sequence can jointly capture absolute positions across video segments and relative dependencies within each segment, thereby mitigating flickering and discontinuities.

\subsection{Direct Forcing}
\label{sec:Direct Forcing}

Autoregressive video generation often suffers from error accumulation caused by the discrepancy between training and inference: during training, the model is conditioned on ground-truth frames, whereas during inference it relies on its own previously generated outputs. 

To mitigate this mismatch, we propose a rectified-flow-based single-step approximation strategy that directly aligns training and inference trajectories while preserving computational efficiency.

\paragraph{Training with Rectified Flow.} 
Following the rectified flow formulation~\cite{rectifiedflow}, we define a linear interpolation between the video distribution $\textbf{x}$ and Gaussian noise $\boldsymbol{\epsilon} \sim \mathcal{N}(0, I)$. For simplicity, we omit the superscript of $\textbf{x}$ and use the subscript to indicate timestep $t$:
\begin{equation}
\textbf{x}_t = t \textbf{x} + (1-t) \boldsymbol{\epsilon}, \quad t \in [0,1].
\label{equ:flow_definition2}
\end{equation}
The instantaneous velocity along this trajectory is given by
\begin{equation}
\boldsymbol{u}_t = \frac{d\textbf{x}_t}{dt} = \textbf{x} - \boldsymbol{\epsilon},
\end{equation}
which defines an ordinary differential equation (ODE) guiding $\textbf{x}_t$ toward the target $\textbf{x}$. The model predicts a velocity field $v_\theta(\textbf{x}_t, t)$, and parameters are optimized by minimizing the flow matching loss:
\begin{equation}
\mathcal{L}_{\mathrm{FM}}(\theta) = \mathbb{E}_{t,\textbf{x},\boldsymbol{\epsilon}} \big[ \lVert v_\theta(\textbf{x}_t, t) - \boldsymbol{u}_t \rVert^2 \big].
\label{equ:FM-loss2}
\end{equation}

\begin{figure}[t] % [t] 表示放在页面顶部
    \centering
    \includegraphics[width=0.8\textwidth]{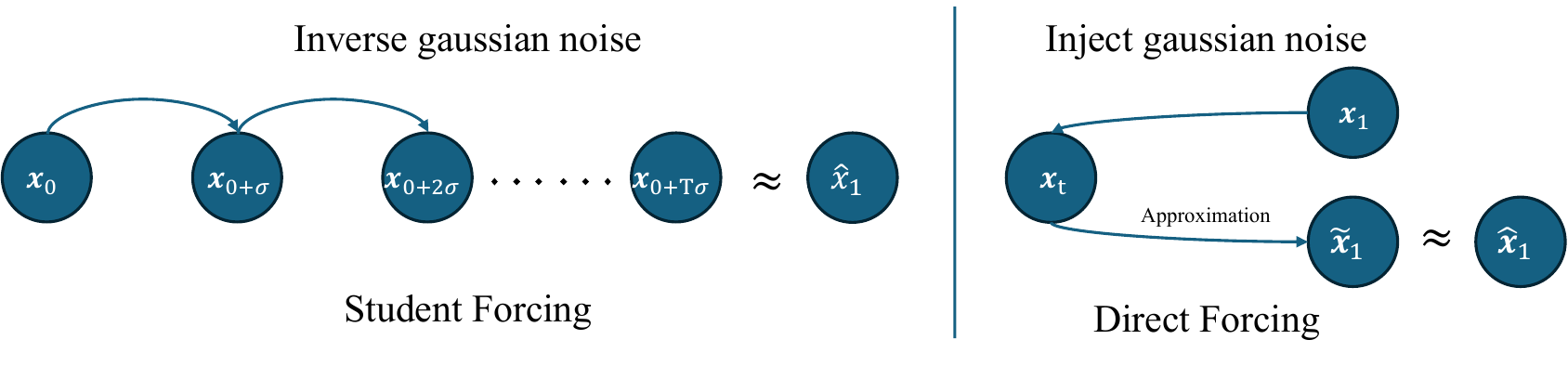}
    \caption{Schematic illustration of the approximation process. In Student Forcing, multi-step inference is applied to approximate $\mathbf{\hat{x}}_1$, but this incurs substantial computational overhead and slows training convergence. In contrast, Direct Forcing applies a single-step transformation from $\mathbf{x}_1$ to $\mathbf{x}_t$, followed by a denoising step that produces $\tilde{\mathbf{x}}_1$ as an estimate of $\mathbf{\hat{x}}_1$. This approach incurs no additional computational burden, thereby enabling faster training.}

    \label{fig:DirectForcing}
\end{figure}
\paragraph{Single-Step Approximation.}
During inference, a video is generated by reverse-time integration of the ODE starting from $\textbf{x}_0 \sim \mathcal{N}(0,I)$ to generated $\hat{\textbf{x}}_1$:
\begin{equation}
\hat{\textbf{x}}_1 = \int_0^1 v_\theta(\textbf{x}_t, t) dt.
\label{inference}
\end{equation}
As illustrated in Fig.~\ref{fig:DirectForcing}, Student Forcing~\citep{studenforcing}, through multi-step inference in Eq.~\ref{inference} as an approximation of $\hat{\textbf{x}}_1$, would incur significant computational costs. Based on Eq.~\ref{equ:single_step}, we align training with inference without incurring high computational costs by approximating the trajectory in a single step:
\begin{equation}
\tilde{\textbf{x}}_1 = \textbf{x}_t + \Delta_{t} *v_\theta(\textbf{x}_t,t) \approx \hat{\textbf{x}}_1,\quad \Delta_t=1-t
\label{equ:single_step}
\end{equation}
Intuitively, since rectified flow guarantees a more direct ODE trajectory, $\textbf{x}_t + \Delta_t v_\theta(\textbf{x}_t, t)$ serves as an effective single-step approximation of the generated distribution, thereby bridging the gap between training and inference. Concretely, the model first uses the ground-truth data $\textbf{x}^{i-1}$ and Eq.~\ref{equ:single_step} to obtain a one-step approximation $\tilde{\textbf{x}}^{i-1}$. This approximation then serves as the conditional input for generating $\textbf{x}^i$ during training, thereby exposing the model to inference-like conditions and effectively mitigating distribution mismatch while reducing error accumulation.

\paragraph{Optimization  Strategy}

In practice, we leverage Direct Forcing to sample clips from the same video in chronological order and use them as conditional inputs for iterative training. Unlike prior approaches \citep{ttt,zhao2024apt} that train clips independently, this design explicitly reinforces temporal continuity across segments. However, applying backpropagation with an optimizer update at every step can perturb the learned distribution and impair the consistency of input distributions across clips during training. To mitigate this issue, we adopt gradient accumulation: gradients are aggregated over multiple clips before performing a single parameter update. This strategy stabilizes optimization and, as our experiments show, substantially improves cross-clip consistency and long-range temporal coherence in generated videos.

\section{Experiment}

\subsection{Implementation Details}

% 我们使用 Framepack-F1 作为基础模型，开展图生成视频任务。训练数据主要来自 Mira、Sekai等数据集，总计约 16,000 段视频，覆盖 150 小时的多样场景，其中 Mira 和 Sekai 的单段视频时长最长可达 1 分钟。为保证数据质量，我们对所有视频进行了动态性和切镜过滤。训练在 64 张 96GB GPU 上并行进行，批量大小为 1，总时长约 4 天。优化器采用 AdamW，初始学习率为 $10^{-5}$。整个训练流程分为两个阶段：第一阶段使用 teacher forcing 训练整个网络，以确保模型在早期快速收敛并具备建模长程依赖（long-term dependency）的能力，避免因采样偏差导致的不稳定训练。第二阶段则采用 Tiforce 对输出层进行微调，在保持 backbone 稳定性的同时，对齐训练与推理过程，从而显著缓解自回归生成中的误差累积问题。

We adopt Framepack-F1 as the backbone model for the image-to-video generation task. The training dataset is primarily sourced from Mira~\citep{ju2024miradata} and Sekai~\citep{li2025sekai}, comprising approximately 16{,}000 video clips with a total duration of 150 hours across diverse scenarios. The longest videos in both Mira and Sekai extend up to one minute. To ensure data quality, we apply dynamism and shot-cut filtering to all samples. Training is conducted in parallel on GPU clusters (96GB memory each) with a batch size of 1 for approximately five days. We employ the AdamW optimizer with an initial learning rate of $10^{-5}$. The training procedure consists of two stages. In the first stage, we apply teacher forcing to train the entire network, which accelerates convergence and mitigates instability caused by sampling bias. In the second stage, we only fine-tune the output layer with Direct Forcing, which could stabilize the backbone and align training with inference, thereby substantially reducing error accumulation in autoregressive generation.

\subsection{Evaluation Metrics}
% 遵循已有的评测协议，我们基于六个定量评判指标对生成视频进行评估：(1) \textbf{imaging\_quality}，(2) \textbf{aesthetic\_quality}，(3) \textbf{dynamic\_degree}，(4) \textbf{motion\_smoothness}，(5) \textbf{background\_consistency}，(6) \textbf{subject\_consistency}。此外，我们还采用了人工评估方案。为了区分不同尺度的视频生成能力，我们将视频划分为三个时长区间：1 分钟对应长视频，30 秒对应中长视频，10 秒对应短视频。
Following established evaluation protocols~\citep{huang2023vbench,zheng2025vbench2}, we assess the generated videos using six quantitative metrics: (1) imaging quality, (2) aesthetic quality, (3) dynamic degree, (4) motion smoothness, (5) background consistency, and (6) subject consistency. In addition, we conduct exposure bias and human evaluation to provide complementary subjective assessments. To distinguish the model’s generation capability across different temporal scales, we categorize videos into three duration ranges: short (10 seconds), medium-length (30 seconds), and long (1 minute).

\textbf{Imaging Quality:} This metric captures distortions in generated frames, including over-exposure, noise, and blur. We measure it using the MUSIQ~\citep{ke2021musiq} image quality predictor trained on the SPAQ~\citep{spaq} dataset.

\textbf{Aesthetic Quality:} We evaluate the aesthetic quality of each video frame using the LAION aesthetic predictor~\citep{schuhmann2022laion}, which considers factors such as composition, color richness and harmony, photorealism, naturalness, and overall artistic value.

\textbf{Dynamic Degree:} We employ RAFT~\citep{teed2020raft} to estimate the extent of motion in synthesized videos.

\textbf{Motion Smoothness:} We leverage motion priors from the video frame interpolation model~\citep{li2023amt}, as adapted by VBench, to evaluate motion smoothness.

\textbf{Background Consistency:} Following VBench, we measure the temporal consistency of background scenes by computing CLIP~\citep{clip} feature similarity across frames.

\textbf{Subject Consistency:} We compute DINO~\citep{dino} feature similarity between frames to assess the consistency of a subject’s appearance throughout the sequence.

% 我们follow \cite{framepack},针对生成的长时视频进行评估，We define the start-end contrast ∆M drift for an arbitrary quality metric M as:。

% 为了准确的测量模型的长时生成能力，我们以最后实际模型推理得到的帧为Vend，而非用户得到的视频最后帧。我们认为评估主体应该是模型的能力，所以应该以模型为主体进行外推。
\paragraph{Exposure Bias Metric:} 
Following \cite{Zhang2025PackingIF}, we evaluate long-horizon video generation by defining the start–end contrast, denoted as $\Delta_{\text{drift}}^M$, for an arbitrary quality metric $M$ (e.g., imaging quality, aesthetic quality).
\begin{equation}
\Delta_{\text{drift}}^M(V) = \big| M(V_{\text{start}}) - M(V_{\text{end}}) \big|,
\end{equation}
Where  V is the tested video, $V_{\text{start}}$ t represents the first 15\% of frames, and $V_{end}$ represents the last 15\% of frames.To accurately measure the model’s long-horizon generation capability, we take the last frame generated by the model itself as $V_{\text{end}}$, rather than the final frame of the user-obtained video. Our rationale is that the evaluation should focus on the model’s capability, and thus extrapolation ought to be defined with respect to the model’s own outputs.

\paragraph{Human Assessment:} 
%  我们从A/B测试中收集人类的偏好，我们生成100条视频，随机的打乱次序并且分发，保证人类评估不会受到干扰。我们follow \cite(pack),提供ELO-K32和对应的排名。
We collect human preferences through A/B testing. Specifically, we generate 160 videos, randomly shuffle their order, and distribute them to evaluators to ensure unbiased assessments. Following \cite{Zhang2025PackingIF}, we report ELO-K32 scores along with the corresponding rankings.

\subsection{Generation Performance}

\begin{table*}[t]
    \centering
    \setlength{\abovecaptionskip}{0pt}
    \setlength{\belowcaptionskip}{0pt}
    \resizebox{\linewidth}{!}{
    \renewcommand{\arraystretch}{1.15}
    \begin{tabular}{c|cccccc|cccc|cc}
        \hline
        \multirow{3}{*}{\textbf{Method}} &  
        \multicolumn{6}{c|}{Global Metrics} &
        \multicolumn{4}{c|}{Error Accumulation} &
        \multicolumn{2}{c}{Human Evaluation} \\
        \cline{2-13}
        & \begin{tabular}{c} Imaging \\ Quality $\uparrow$ \end{tabular} &
          \begin{tabular}{c} Aesthetic \\ Quality $\uparrow$ \end{tabular} &
          \begin{tabular}{c} Dynamic \\ Degree $\uparrow$ \end{tabular} &
          \begin{tabular}{c} Motion \\ Smoothness $\uparrow$ \end{tabular} &
          \begin{tabular}{c} Background \\ Consistency $\uparrow$ \end{tabular} &
          \begin{tabular}{c} Subject \\ Consistency $\uparrow$ \end{tabular} &
          \begin{tabular}{c} $\Delta$Imaging \\ Quality $\downarrow$ \end{tabular} &
          \begin{tabular}{c} $\Delta$Aesthetic \\ Quality $\downarrow$ \end{tabular} &
          \begin{tabular}{c} $\Delta$Background \\ Consistency $\downarrow$ \end{tabular} &
          \begin{tabular}{c} $\Delta$Subject \\ Consistency $\downarrow$ \end{tabular} &
          \begin{tabular}{c} ELO \\ $\uparrow$ \end{tabular} &
          \begin{tabular}{c} Rank \\ $\downarrow$ \end{tabular} \\
        \hline
        Magi-1  & 54.64\% & 53.98\% & \textbf{67.5\%} & 99.17\% & 89.30\% & 82.33\%& 4.19\% & 5.02\% & 1.53\%& 0.97\% & 1434 & 4\\
        FramePack-F0  & \textbf{68.06\%}& \textbf{62.89\%}& 15.38\%& 99.27\%& 92.22\%& 90.03\%& \textbf{2.34\%} & \textbf{2.45\%}& 2.68\%& 2.99\%& 1459 & 3\\
        FramePack-F1  & 67.06\%& 59.11\%& 53.85\%& 99.12\% & 90.21\% & 83.48\% & 2.71\%& 4.57\%& 1.59\%& 1.08\%& 1537 & 2 \\
        \hline
        Ours & 67.55\% & 59.75\% & 48.37\% & \textbf{99.31\%} & \textbf{93.25\%} & \textbf{91.16\%} & 2.51\% & 3.25\%& \textbf{1.21\%}& \textbf{0.76\%}& \textbf{1568} &\textbf{1}\\
        \hline
    \end{tabular}
    }
    \vspace{1ex}
    \caption{Comparison of different methods on \textbf{Global Metrics}, \textbf{Error Accumulation Metrics}, and \textbf{Human Evaluation}. 
    The best results are highlighted in \textbf{bold}.}

    \label{tab:comparison}
\end{table*}

\begin{figure}[t]
    \centering
    \setlength{\tabcolsep}{0pt}
    \begin{tabular}{m{0.04\textwidth} m{0.95\textwidth}}
        & \makebox[0.95\textwidth][c]{%
            \makebox[0.19\textwidth]{0s}%
            \makebox[0.19\textwidth]{7.5s}%
            \makebox[0.19\textwidth]{15s}%
            \makebox[0.19\textwidth]{22.5s}%
            \makebox[0.19\textwidth]{30s}%
        } \\[0.3em]

        \centering\rotatebox{90}{mage} &
        \includegraphics[width=\linewidth,height=0.25\textheight,keepaspectratio,page=1]{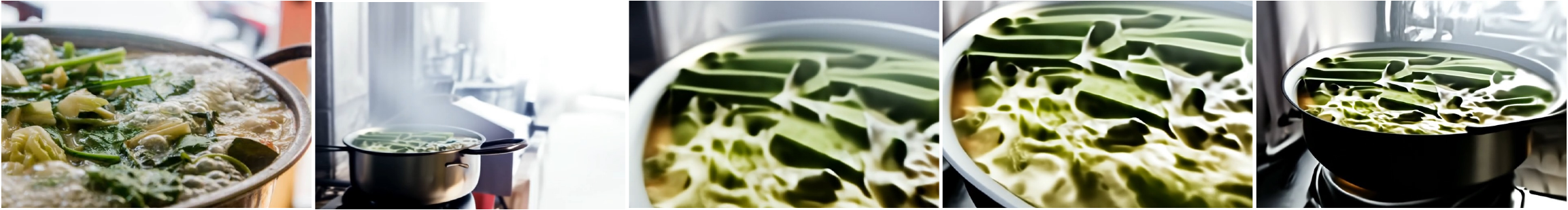} \\[0.1em]

        \centering\rotatebox{90}{f0} &
        \includegraphics[width=\linewidth,height=0.25\textheight,keepaspectratio,page=1]{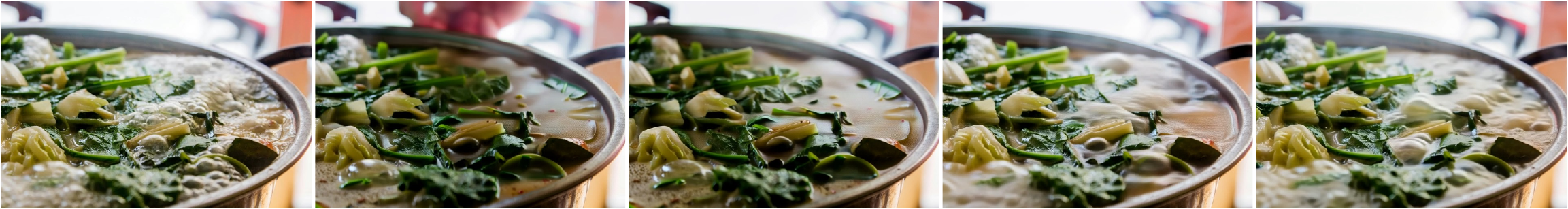} \\[0.1em]

        \centering\rotatebox{90}{f1} &
        \includegraphics[width=\linewidth,height=0.25\textheight,keepaspectratio,page=1]{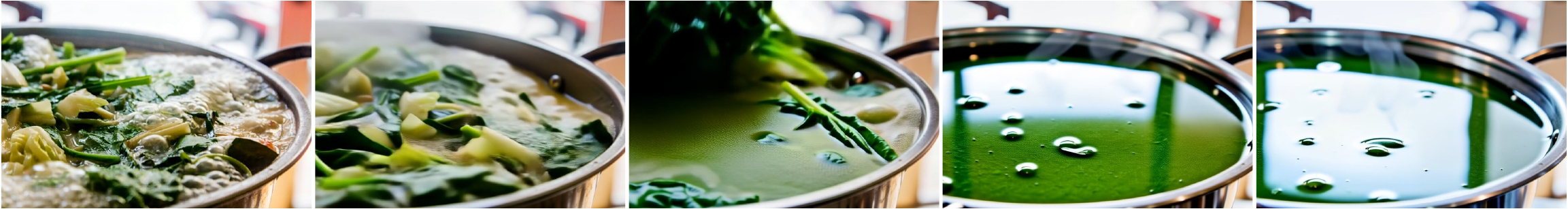} \\[0.1em]

        \centering\rotatebox{90}{ours} &
        \includegraphics[width=\linewidth,height=0.25\textheight,keepaspectratio,page=1]{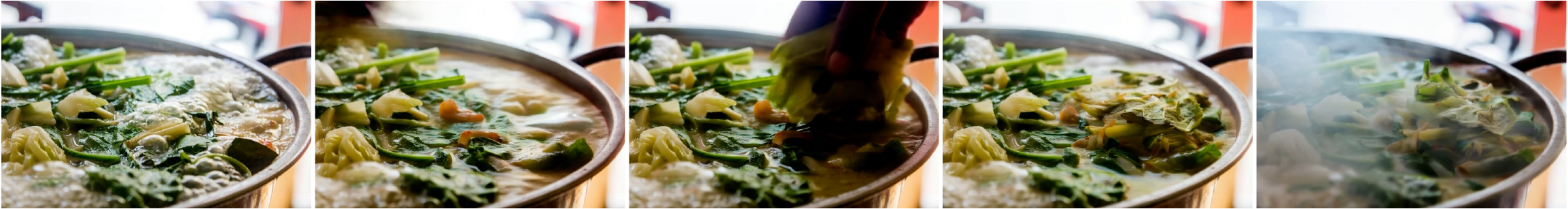} \\
    \end{tabular}
    
    \caption{Visualization of 30-second videos comparing all methods in terms of consistency preservation and interaction capability. \textbf{Prompt:} Close-up view of vegetables being added into a large silver pot of simmering broth, with leafy greens and stems swirling vividly in the bubbling liquid. Rising steam conveys warmth and motion, while blurred kitchen elements and natural light in the background create a homely yet dynamic culinary atmosphere.}
    \label{pic:compare1}
\end{figure}

\begin{figure}[t]
    \centering
    \setlength{\tabcolsep}{0pt}
    \begin{tabular}{m{0.04\textwidth} m{0.95\textwidth}}
        & \makebox[0.95\textwidth][c]{%
            \makebox[0.19\textwidth]{0s}%
            \makebox[0.19\textwidth]{15s}%
            \makebox[0.19\textwidth]{30s}%
            \makebox[0.19\textwidth]{45s}%
            \makebox[0.19\textwidth]{60s}%
        } \\[0.3em]
        \centering\rotatebox{90}{mage} &
        \includegraphics[width=\linewidth,height=0.25\textheight,keepaspectratio,page=1]{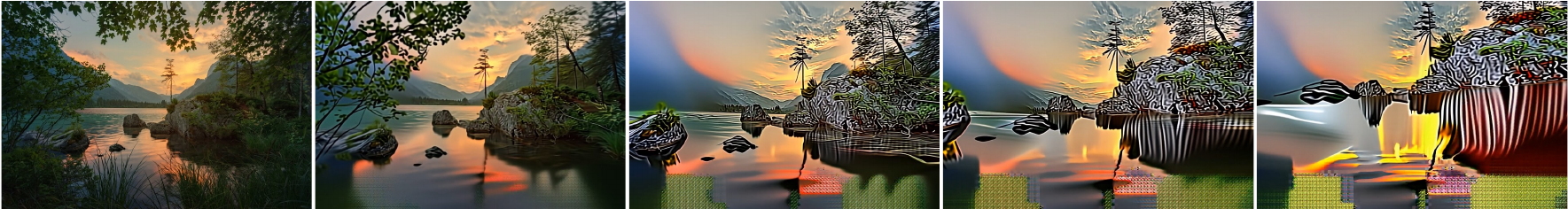} \\[0.1em]

        \centering\rotatebox{90}{f0} &
        \includegraphics[width=\linewidth,height=0.25\textheight,keepaspectratio,page=1]{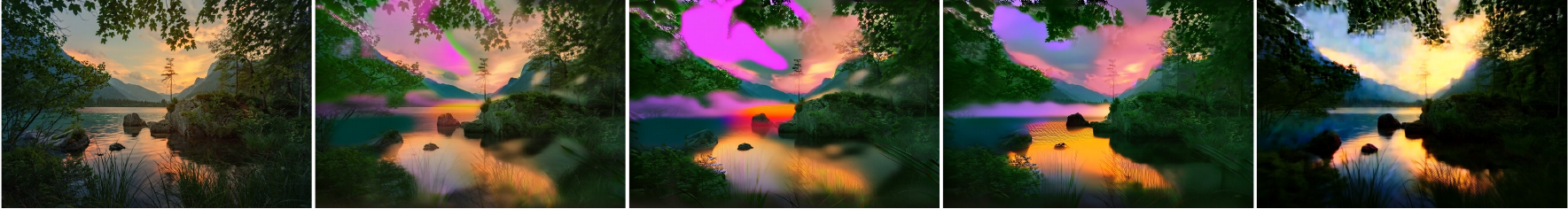} \\[0.1em]

        \centering\rotatebox{90}{f1} &
        \includegraphics[width=\linewidth,height=0.25\textheight,keepaspectratio,page=1]{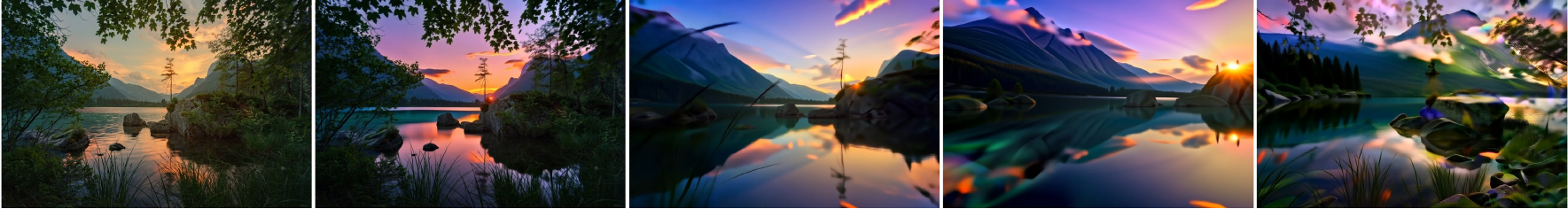} \\[0.1em]

        \centering\rotatebox{90}{ours} &
        \includegraphics[width=\linewidth,height=0.25\textheight,keepaspectratio,page=1]{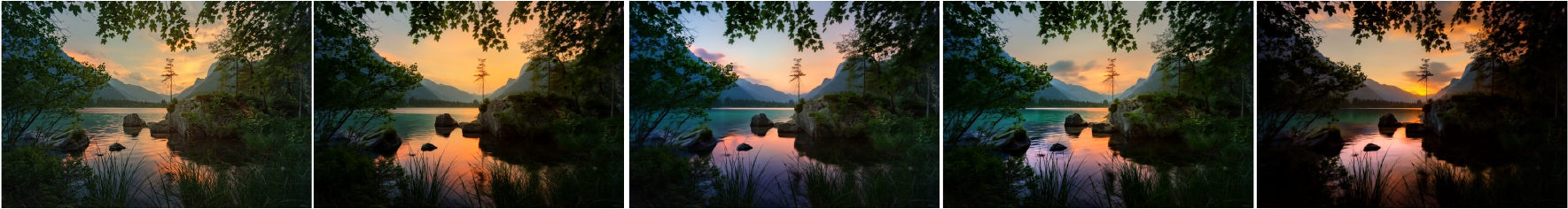} \\
    \end{tabular}
    \caption{Visualization of a 60-second video illustrating the accumulation of errors. Our method maintains image quality comparable to the first frame even over minute-long sequences. \textbf{Prompt:} The sun sets over a serene lake nestled within majestic mountains, casting a warm, golden glow that softens at the horizon. The sky is a vibrant canvas of orange, pink, and purple, with wispy clouds catching the last light. Calm and reflective, the lake's surface mirrors the breathtaking colors of the sky in a symphony of light and shadow. In the foreground, lush greenery and rugged rocks frame the tranquil scene, adding a sense of life and stillness. Majestic, misty mountains rise in the background, creating an overall atmosphere of profound peace and tranquility.}

    \label{pic:compare2}
\end{figure}

\begin{figure}[t]
    \centering
    \setlength{\tabcolsep}{0pt}
    \begin{tabular}{m{0.04\textwidth} m{0.95\textwidth}}    
        \\[0.3em]
        \centering\rotatebox{90}{mage} &
        \includegraphics[width=\linewidth,height=0.25\textheight,keepaspectratio,page=1]{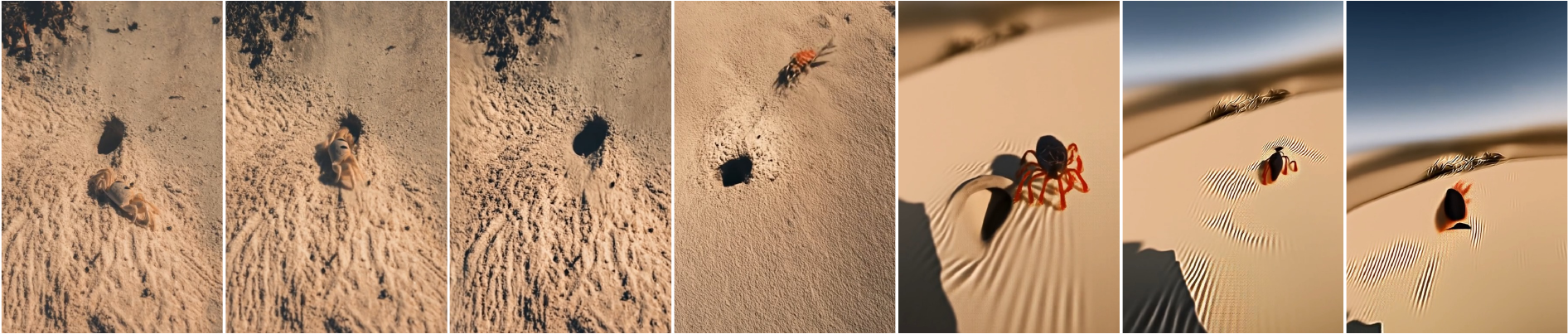} \\[0.1em]

        \centering\rotatebox{90}{f0} &
        \includegraphics[width=\linewidth,height=0.25\textheight,keepaspectratio,page=1]{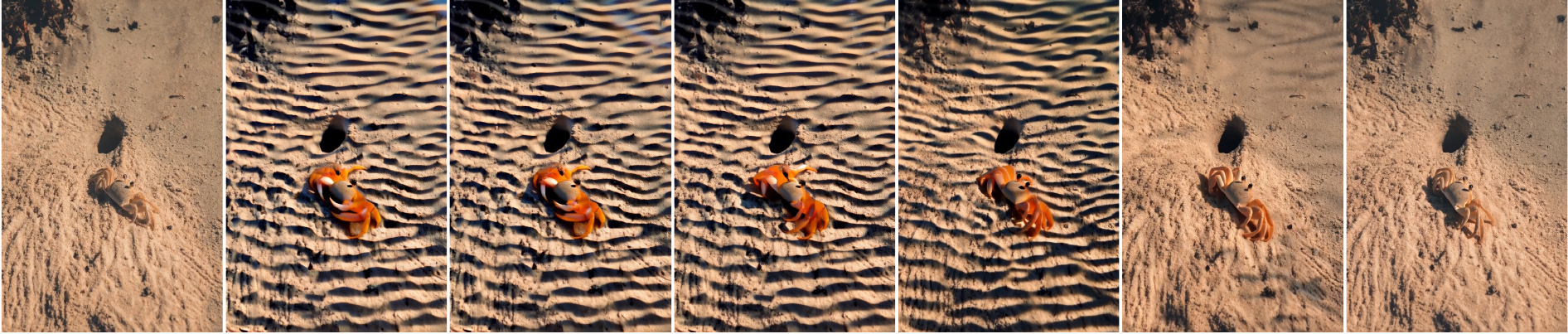} \\[0.1em]

        \centering\rotatebox{90}{f1} &
        \includegraphics[width=\linewidth,height=0.25\textheight,keepaspectratio,page=1]{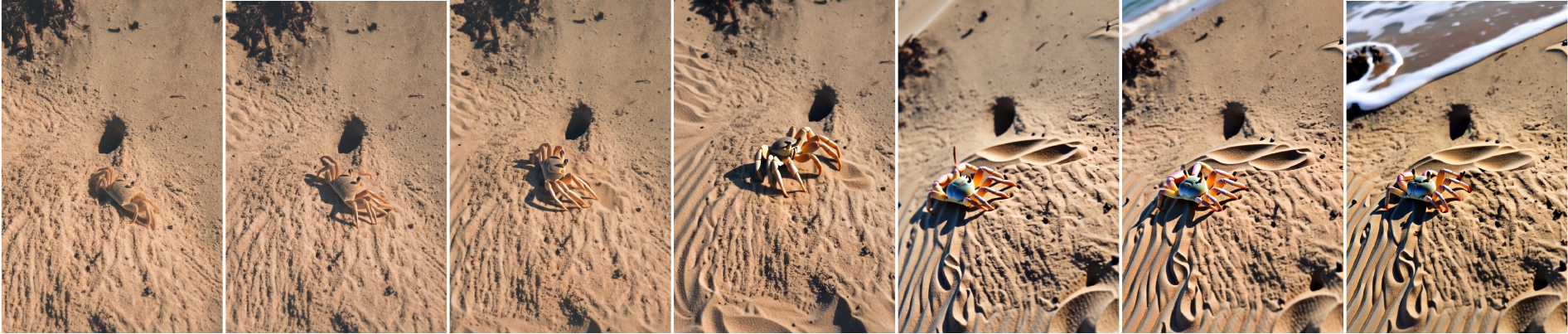} \\[0.1em]

        \centering\rotatebox{90}{ours} &
        \includegraphics[width=\linewidth,height=0.25\textheight,keepaspectratio,page=1]{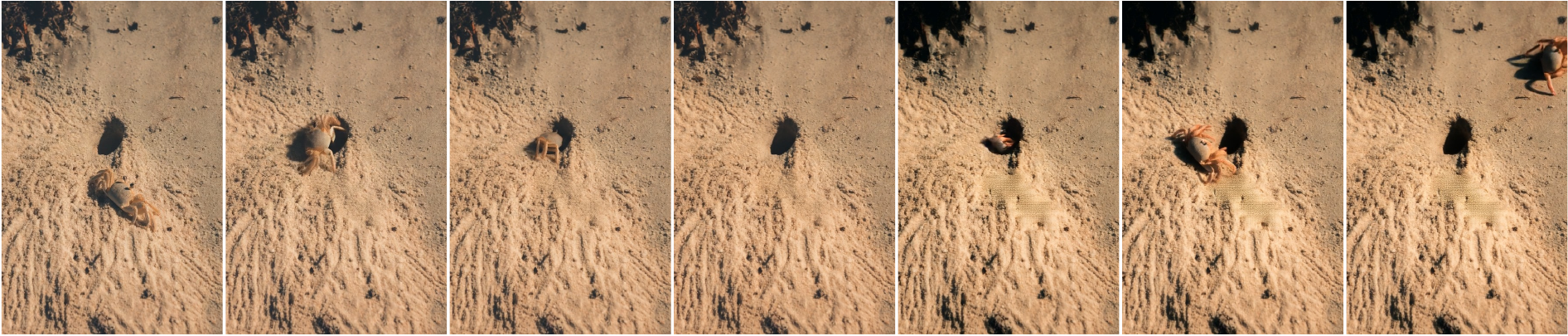} \\
    \end{tabular}
\caption{Consistency evaluation on a 60-second video shows that when an object ID is heavily occluded for an extended period, reconstruction remains challenging. Both F0 and F1 fail to follow the prompt and exhibit noticeable error accumulation. Although MAGI-1 follows the prompt, it is unable to maintain temporal consistency. \textbf{Prompt:} On the peaceful, sun-drenched sandy beach, a small crab first retreats into its burrow before reemerging. The lens captures its shimmering shell and discreet stride under the low sun angle. As it slowly crawls outward, the crab leaves a faint trail behind, while its elongated shadow adds a cinematic texture to this tranquil scene.}
    \label{pic:consist}
\end{figure}

% 我们选择FramePack-F0，FramePack-F1和Magi-1作为我们的基准线。我们总共测试a个10秒视频，b个30秒视频，c个1分钟视频，分辨率都是480p,24fps. 
We compare our method with FramePack-F0, FramePack-F1~\citep{Zhang2025PackingIF}, and Magi-1~\citep{ai2025MAGI1AV}. Evaluation is conducted on 160 videos: 60 of 10 seconds, 60 of 30 seconds, and 40 of 60 seconds, all at 480p and 24 fps. All images are sourced from Vbench~\citep{huang2024vbench++}, and prompts are rewritten using Qwen2.5-Vl~\citep{Qwen2.5-VL}. Quantitative results are reported in Table~\ref{tab:comparison} and Table~\ref{tab:ablation}, while qualitative comparisons are shown in Fig.~\ref{pic:compare1}, Fig.~\ref{pic:compare2}, and Fig.~\ref{pic:consist}.

\paragraph{Qualitative results.}
Representative 30-second and 60-second examples are shown in Fig.~\ref{pic:compare1} and Fig.~\ref{pic:compare2}. For brevity, we simplify the prompts associated with generated videos. Our method exhibits fewer temporal identity shifts and geometric distortions compared to FramePack-F1 and Magi-1. While FramePack-F0 preserves visual fidelity in Fig.~\ref{pic:compare1}, it demonstrates reduced inter-frame dynamics, consistent with its quantitative metrics. Notably, as video length increases, competing methods suffer from more severe error accumulation, whereas our approach maintains video quality, including aesthetics and character consistency. These results highlight the effectiveness of Direct Forcing in enhancing long-term video generation.

\paragraph{Quantitative results.}
As reported in Table~\ref{tab:comparison}, our method achieves the best performance on Background Consistency, Subject Consistency, and Motion Smoothness, demonstrating a strong ability to preserve long-term temporal coherence and generate smooth motion. While FramePack-F0 attains higher Image Quality and Aesthetic Quality, it exhibits weaker inter-frame dynamics, which we attribute to its anti-drift sampling strategy. Conversely, Magi-1 produces stronger dynamics but suffers from degraded temporal and subject consistency. In human evaluation, our method with ELO-32K still achieves the best overall performance. These findings are further corroborated by the qualitative results in Fig.~\ref{pic:compare1} and Fig.~\ref{pic:compare2}. Importantly, our approach also achieves the lowest Error Accumulation, underscoring its stability in long-term video generation.

\paragraph{Ablation study.}
As reported in Table~\ref{tab:ablation}, all experimental results were obtained using a subset of the training data to conserve computational resources. To assess the influence of the training dataset, we first fine-tuned FramePack-F1 (F1 with CT) on our dataset. This slightly improved the Dynamic Degree metric but degraded performance on other evaluations, demonstrating the effectiveness of our proposed MemoryPack and Direct Forcing. 

To further examine the semantic contribution of MemoryPack, we initialized the global memory $\psi_0$ with a zero vector (zero-MemoryPack). This led to worse performance on error-accumulation metrics, which we attribute to the absence of semantic guidance, resulting in reduced consistency. These findings indicate that semantic guidance stabilizes long-term video generation. 

We also ablated Direct Forcing by training the model with its actual sampling process as input. To balance training efficiency, we set the sampling step to 5; however, the resulting performance remained substantially inferior to Direct Forcing, underscoring its effectiveness for long-term video generation.

\paragraph{Consistency visualization.}
Figure~\ref{pic:consist} illustrates object reconstruction after prolonged disappearance. Over a 60-second sequence, our method accurately reconstructs objects that remain absent for extended periods. Notably, even when subjects temporarily vanish due to occlusion, the model reconstructs and generates objects with consistent identity and 2D structure after long intervals. These results demonstrate that MemoryPack effectively preserves long-term contextual information, enabling stable memory for extended video generation.
\begin{table*}[t]
    \centering
    \vspace{2ex}
    \resizebox{\linewidth}{!}
    {
    \renewcommand{\arraystretch}{1.15}
    \begin{tabular}{c|cccccc|cccc}
        \hline
        \multirow{2}{*}{\textbf{Method}} &  
        \multicolumn{6}{c|}{Global Metrics} &
        \multicolumn{4}{c}{Error Accumulation} \\
        \cline{2-11}
        & \begin{tabular}{c} Imaging \\ Quality $\uparrow$ \end{tabular} &
          \begin{tabular}{c} Aesthetic \\ Quality $\uparrow$ \end{tabular} &
          \begin{tabular}{c} Dynamic \\ Degree $\uparrow$ \end{tabular} &
          \begin{tabular}{c} Motion \\ Smoothness $\uparrow$ \end{tabular} &
          \begin{tabular}{c} Background \\ Consistency $\uparrow$ \end{tabular} &
          \begin{tabular}{c} Subject \\ Consistency $\uparrow$ \end{tabular} &
          \begin{tabular}{c} $\Delta$Imaging \\ Quality $\downarrow$ \end{tabular} &
          \begin{tabular}{c} $\Delta$Aesthetic \\ Quality $\downarrow$ \end{tabular} &
          \begin{tabular}{c} $\Delta$Background \\ Consistency $\downarrow$ \end{tabular} &
          \begin{tabular}{c} $\Delta$Subject \\ Consistency $\downarrow$ \end{tabular} \\
        \hline
       F1 w/ CT &  54.33\% & 53.07\% & 56.77\% & 98.81\% & 85.32\% & 85.64\% & 4.32\% & 7.18\% & 3.22\% & 1.32\% \\
       MemoryPack & 55.31\% & \textbf{60.69}\% & 51.13\% & 98.86\% & \textbf{88.21\%} & 86.77\% & \textbf{2.47}\% & 4.99\% & 2.31\% & 1.88\% \\
       zero-MemoryPack & 57.37\%  & 51.91\% & \textbf{62.71\%} & 98.85\% & 87.31\% & 86.21\% & 8.32\% & 6.31\% & 2.43\% & 3.10\% \\
       Student forcing & 49.32\% & 52.33\% & 59.88\% & 98.96\% & 85.32\% & 82.41\% & 3.17\% & 5.11\% & \textbf{2.02\%} & 1.12\% \\
       All & \textbf{57.65}\% & 55.75\% & 55.37\% & \textbf{99.11\%} & 87.17\% & \textbf{88.77\%} & 3.21\% & \textbf{4.77\%} & \textbf{2.02\%} & \textbf{0.99\%} \\
        \hline 
    \end{tabular}
    }
    \caption{Ablation Study on Different Model Structures and Strategies.}
    \label{tab:ablation}
\end{table*}
\section{Conclusion}

% 我们提出的MemoryPack是一个没有额外3D先验或固定规则选择这样的显式启发式的情况下，只依赖视频帧和文本成功捕捉到长远期的记忆信息，我们证明无需额外增加大量的计算资源，和改变核心的Dit架构通过轻量式的模型架构即可从历史帧数据中学习长远期信息。同时Tiforce提出一种新的方式，缓和自回归视频生成人物的训推不一致问题，实验结果证明，通过Tiforce能够有效的避免错误累积，从而加强ip一致性和缓和曝光偏差。我们的方法为下一代，长视频生成模型提供了解决思路。
We propose MemoryPack, a lightweight mechanism that jointly models long- and short-term memory from visual inputs and text prompts, enabling the learning of long-range dependencies without requiring 3D priors, handcrafted heuristics, or modifications to the core DiT framework. We further introduce Direct Forcing, a simple yet effective approach for mitigating the training–inference discrepancy in autoregressive video generation, which incurs no additional training cost or computational overhead. Experimental results show that MemoryPack and Direct Forcing reduce error accumulation, enhance identity-preservation consistency, and alleviate exposure bias. Together, these contributions pave the way for next-generation long-form video generation models.

\paragraph{Limitation :} This work explores MemoryPack for modeling both long- and short-term memory. In contrast, SemanticPack employs a sliding-window Transformer backbone to capture long-term information. This design achieves linear computational complexity and enables efficient training and inference. However, it still introduces artifacts in highly dynamic scenarios, and its ability to maintain long-term consistency in hour-long videos remains limited. Similarly, Direct Forcing adopts a simple and efficient single-step approximation strategy, yet its effectiveness is highly dependent on the pre-trained model. As a result, training currently requires multiple stages, and whether this fitting strategy can be integrated into a single-stage pipeline remains an open question. We leave these directions for future work.

\section{Ethics Statement}
This work focuses on algorithmic advances in long-form video generation. Our study does not involve sensitive personal data, or copyrighted materials. The datasets used (e.g., \cite{li2025sekai,ju2024miradata}) are publicly available benchmarks that comply with community standards.  our research is centered on methodological improvements in temporal coherence and computational efficiency. We encourage responsible use of the proposed techniques and recommend safeguards such as watermarking and content moderation in downstream applications.

\section{Reproducibility Statement}
We have made substantial efforts to ensure the reproducibility of our work. Detailed descriptions of the proposed MemoryPack and Direct Forcing mechanisms are provided in Sections~\ref{sec:MemoryPack} and \ref{sec:Direct Forcing}, respectively. To further support reproducibility, we will release source code, configuration files, and pretrained checkpoints upon publication. These resources will enable researchers to replicate our experiments under the same settings and adapt the framework to new datasets or tasks with minimal modifications. Comprehensive documentation and usage guidelines will also be provided to lower the barrier to reproduction and extension. We hope these efforts will not only validate our findings but also foster future research in long-term video generation.

\bibliographystyle{plainnat} 
\bibliography{neurips_2025}
\newpage
\appendix

\section{Appendix}
We provide additional experimental details and visualizations in this section. In Sec.~\ref{app-section1}, we present the optional structures of SemanticPack, where we conduct both qualitative and quantitative evaluations to analyze their impact on modeling efficiency, representation capacity, and overall consistency performance. Sec.~\ref{app-section2} describes the training setup of Direct Forcing, including implementation details, hyperparameter configurations, and training dynamics, to facilitate reproducibility and provide deeper insight into its effectiveness. In Sec.~\ref{app-section3}, we provide consistency visualizations, including minute-level examples, to evaluate the model’s ability to reconstruct historical information under long-term sequences and severe occlusions, while preserving spatial layouts, maintaining subject fidelity, and mitigating temporal drifting.

\subsection{Validation of SemanticPack Models}
\label{app-section1}
\begin{figure}[h!]
    \centering
    \begin{subfigure}[b]{0.32\textwidth} % 宽度可以根据需要微调，例如 0.31, 0.32
        \centering
        \includegraphics[width=\linewidth, page=1]{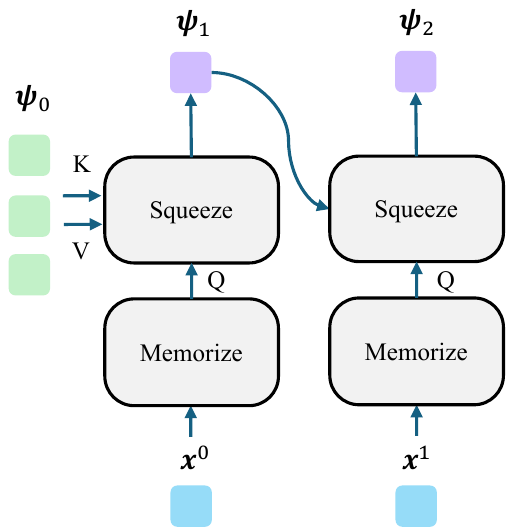}
        \caption{Alternative structure A.}
        \label{fig:pdf1}
    \end{subfigure}% <--- 注意这里的百分号，用来消除换行符可能产生的空格
    \hfill
    \begin{subfigure}[b]{0.32\textwidth}
        \centering
        \includegraphics[width=\linewidth, page=1]{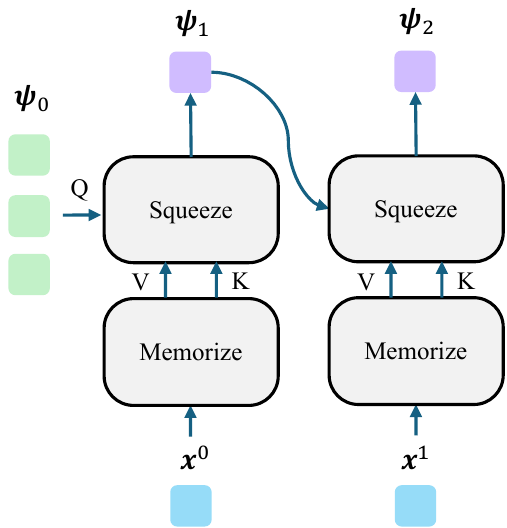}
        \caption{Alternative structure B.}
        \label{fig:pdf2}
    \end{subfigure}% <--- 同样注意这里的百分号
    \hfill
    \begin{subfigure}[b]{0.32\textwidth}
        \centering
        \includegraphics[width=\linewidth, page=1]{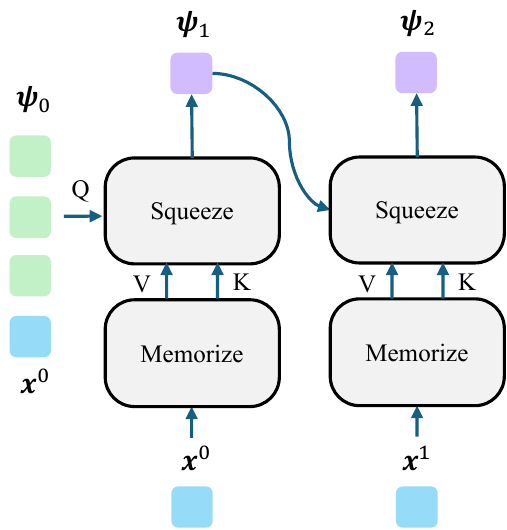}
        \caption{Alternative structure C.}
        \label{fig:pdf3}
    \end{subfigure}

    \caption{Illustration of the optional architecture of SemanticPack.}
    \label{fig:app-se-val}
\end{figure}

We conducted further ablation studies against Squeeze to evaluate the network’s capacity for capturing long-term dependencies. As shown in Fig.~\ref{fig:app-se-val}, we designed three fusion schemes:  
(a) text and image features are used as activation vectors for $K$ and $V$, while the visual representation serves as the query;  
(b) text and image features are used as the query, with the visual representation as key/value;  
(c) to enrich the query, we concatenate text and image features with the visual representation from the first window, and then follow the same setting as (b) for subsequent steps.

In Tab.~\ref{tab:app-sem-1}, we present quantitative results. We observe that structures B and C cause a notable degradation in visual quality and a substantial reduction in temporal dynamics. We attribute these effects to the following reasons: in structure B, the number of text and image tokens is considerably smaller than the number of query tokens in structure A, which is insufficient to capture adequate visual representations, thereby impairing the model's ability to model dynamics. In structure C, although incorporating the initial windows into the query increases the token count, it also introduces multi-modal information, thereby increasing training difficulty.

\begin{table}[h!]
    \centering
    \vspace{2ex}
    \resizebox{\linewidth}{!}
    {
    \renewcommand{\arraystretch}{1.15}
    \begin{tabular}{c|cccccc}
        \hline
        \multirow{2}{*}{\textbf{Method}} & 
        \multicolumn{6}{c}{Global Metrics} \\
        \cline{2-7}
        & \begin{tabular}{c} Imaging \\ Quality $\uparrow$ \end{tabular} &
          \begin{tabular}{c} Aesthetic \\ Quality $\uparrow$ \end{tabular} &
          \begin{tabular}{c} Dynamic \\ Degree $\uparrow$ \end{tabular} &
          \begin{tabular}{c} Motion \\ Smoothness $\uparrow$ \end{tabular} &
          \begin{tabular}{c} Background \\ Consistency $\uparrow$ \end{tabular} &
          \begin{tabular}{c} Subject \\ Consistency $\uparrow$ \end{tabular} \\
        \hline
        C & 48.31\% & 53.15\% & 28.98\% & 97.72\% & 83.27\% & 83.74\% \\
        B & 50.11\% & 50.71\% & 32.78\% & \textbf{98.91}\% & 87.11\% & 80.56\% \\
        A & \textbf{55.31}\% & \textbf{60.69}\% & \textbf{51.13}\% & 98.86\% & \textbf{88.21\%} & \textbf{86.77}\% \\
        \hline 
    \end{tabular}
    }
    \caption{Ablation Study on Different Model Structures and Strategies. Among all variants, structure A achieves the best overall performance.}
    \label{tab:app-sem-1}
\end{table}

\begin{figure}[t]
    \centering
    \setlength{\tabcolsep}{0pt}
    \begin{tabular}{m{0.04\textwidth} m{0.95\textwidth}}
        % --- Timestamp Row ---
        & 
        % This line now automatically calculates the width for each of the 6 timestamps
        % to ensure they are perfectly aligned with the image frames below.
        \makebox[\dimexpr\linewidth/6\relax][c]{0s}%
        \makebox[\dimexpr\linewidth/6\relax][c]{2s}%
        \makebox[\dimexpr\linewidth/6\relax][c]{4s}%
        \makebox[\dimexpr\linewidth/6\relax][c]{6s}%
        \makebox[\dimexpr\linewidth/6\relax][c]{8s}%
        \makebox[\dimexpr\linewidth/6\relax][c]{10s}
        \\[0.3em]

        % --- Image Rows ---
        \centering\rotatebox{90}{C} &
        \includegraphics[width=\linewidth,page=1]{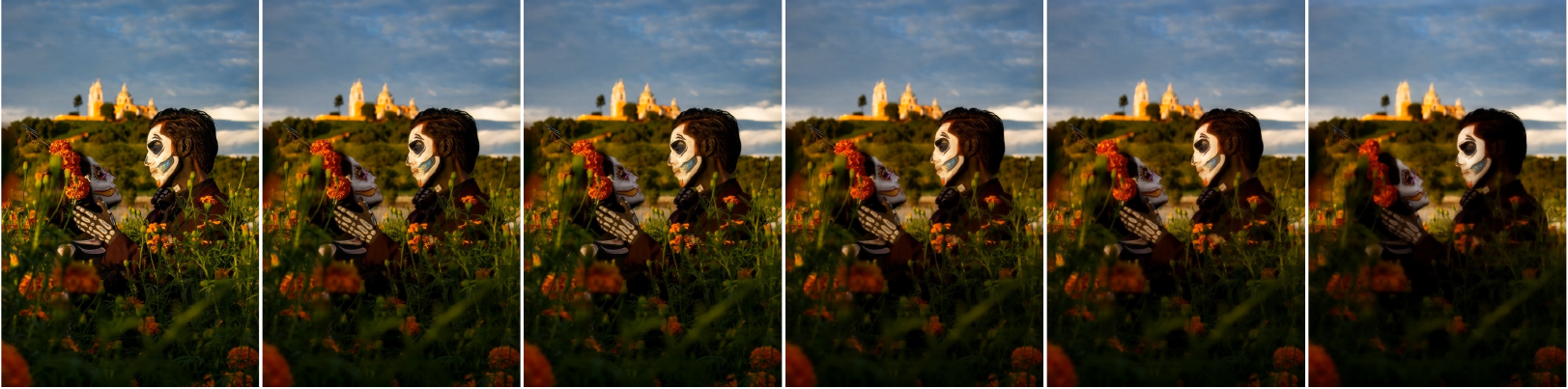} \\[0.1em]
        \centering\rotatebox{90}{B} &
        \includegraphics[width=\linewidth,page=1]{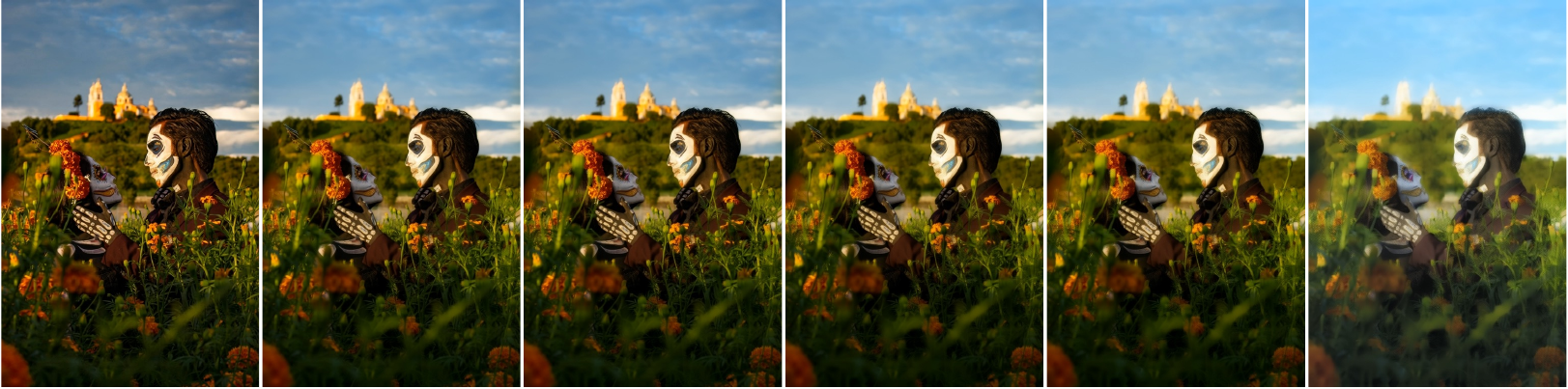} \\[0.1em]
        \centering\rotatebox{90}{A} &
        \includegraphics[width=\linewidth,page=1]{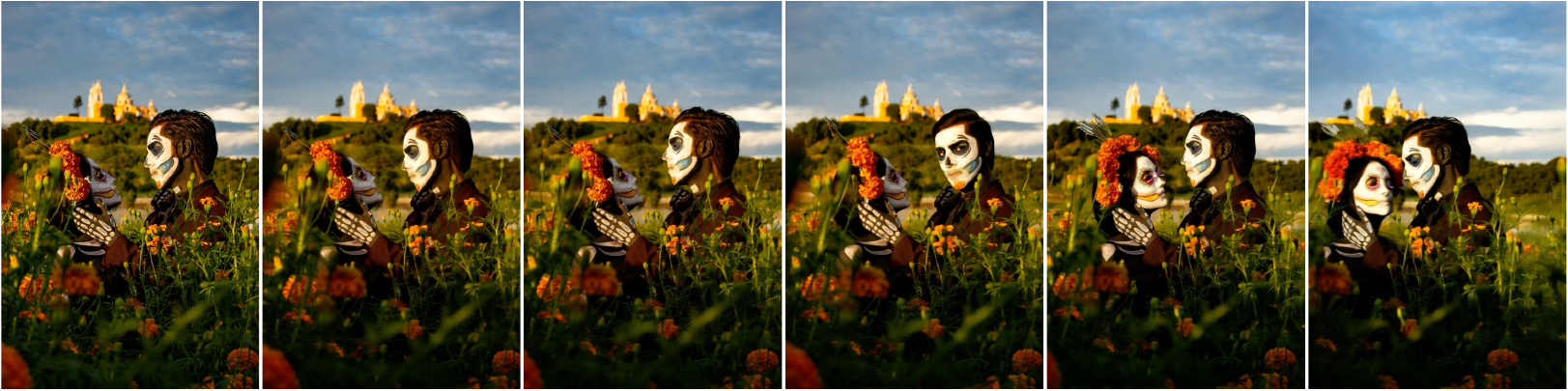} \\[0.1em]

    \end{tabular}
    \caption{\textbf{Prompt:} In a field of golden marigolds, a man and woman stood entwined, their faces glowing with intricate sugar skull makeup beneath the setting sun. The woman, crowned with fiery orange blossoms, gazed at him with tender devotion. He met her eyes, the bold black-and-white patterns on his face striking against his chestnut jacket, his hands gently interlaced with hers. Turning briefly toward the camera, he then lowered his head to kiss her. Behind them, hilltops crowned with golden-domed buildings shimmered beneath a sky of soft blues and pinks, completing the serene, magical scene.}

    \label{pic:app-sem-1}
    \vspace{-2ex}
\end{figure}

To further validate our conclusions, we conduct visualization experiments for the three structures presented in Fig.~\ref{pic:app-sem-1}. The results reveal clear differences in temporal consistency: while structure~A maintains relatively stable motion and coherent frame-to-frame transitions, structures~B and~C exhibit a pronounced degradation in temporal dynamics. In particular, both B and C suffer from noticeable  blurred object boundaries over time, indicating their limited capacity to preserve long-range dependencies. Moreover, they accumulate substantially higher exposure bias compared to structure~A, which further amplifies temporal drift and undermines overall video fidelity.

\subsection{Direct Forcing Training Details}
\label{app-section2}
For Direct Forcing, we employ a data sampler to standardize video clips and ensure sequential training. Since all clips are accumulated through gradient accumulation, the effective number of optimization steps becomes limited. To mitigate this issue, we adopt a curriculum learning strategy by sorting videos according to their length. Specifically, we begin training on videos containing a single clip, which effectively reduces the task to an image-to-video setting and lowers the training difficulty. We then progressively increase the number of accumulated clips, enabling the model to gradually align training with inference, while naturally extending to the autoregressive generation paradigm. This staged progression effectively enhances training stability. In terms of trainable parameters, we restrict updates to the final normalization layers and the output linear layer. Our rationale is that the discrepancy between training and inference primarily arises from a small number of accumulated errors, which eventually lead to drift; hence, adapting only the final layers is sufficient to correct this mismatch.

\subsection{ Visulization}
\begin{figure*}[t]
    \centering
    % We use \setlength to remove any horizontal space before the content.
    \setlength{\tabcolsep}{0pt}
    \begin{tabular}{m{0.04\textwidth} m{0.95\textwidth}}
        & 
        % Each \makebox is 1/7th of the line width, centering the timestamp over its frame.
        % The '%' at the end of each line is crucial to prevent unwanted spaces between the boxes.
        \makebox[\dimexpr\linewidth/6\relax][c]{0s}%
        \makebox[\dimexpr\linewidth/6\relax][c]{1s}%
        \makebox[\dimexpr\linewidth/6\relax][c]{2s}%
        \makebox[\dimexpr\linewidth/6\relax][c]{3s}%
        \makebox[\dimexpr\linewidth/6\relax][c]{4s}%
        \makebox[\dimexpr\linewidth/6\relax][c]{5s}
        \\[0.3em]

         &
        \includegraphics[width=\linewidth,page=1]{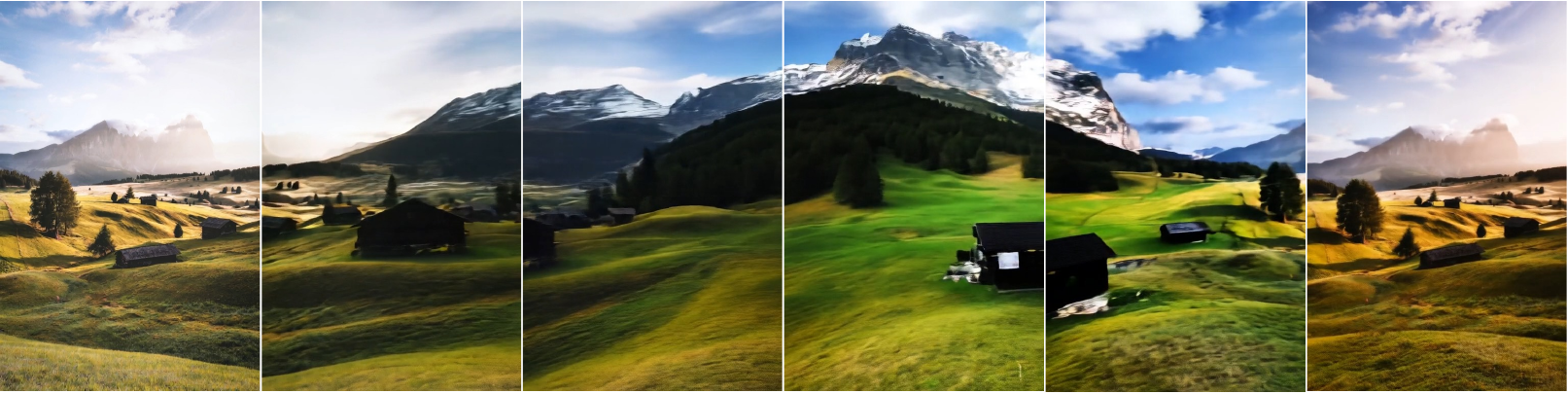} \\[0.1em]

        &
        \makebox[\dimexpr\linewidth/6\relax][c]{0s}%
        \makebox[\dimexpr\linewidth/6\relax][c]{6s}%
        \makebox[\dimexpr\linewidth/6\relax][c]{12s}%
        \makebox[\dimexpr\linewidth/6\relax][c]{18s}%
        \makebox[\dimexpr\linewidth/6\relax][c]{24s}%
        \makebox[\dimexpr\linewidth/6\relax][c]{30s}%
        \\[0.3em]
        
         &
        \includegraphics[width=\linewidth,page=1]{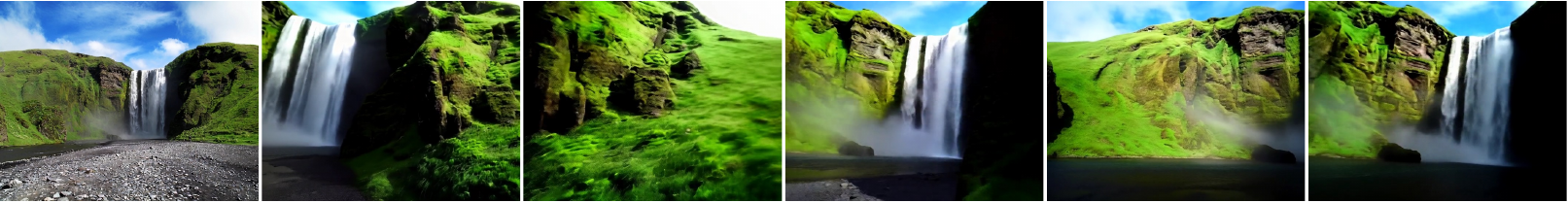} \\[0.1em]

        &
        \makebox[\dimexpr\linewidth/6\relax][c]{0s}%
        \makebox[\dimexpr\linewidth/6\relax][c]{6s}%
        \makebox[\dimexpr\linewidth/6\relax][c]{12s}%
        \makebox[\dimexpr\linewidth/6\relax][c]{18s}%
        \makebox[\dimexpr\linewidth/6\relax][c]{24s}%
        \makebox[\dimexpr\linewidth/6\relax][c]{30s}
        \\[0.3em]
        
         &
        \includegraphics[width=\linewidth,page=1]{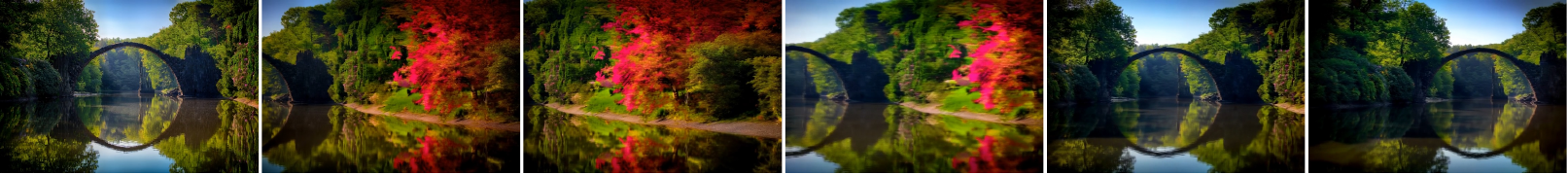} \\[0.1em]

    \end{tabular}
    \caption{Visualization of long-term consistency. The model is evaluated on videos of 5 s and 30 s to assess its ability to maintain coherence.}

    \label{pic:app-sem-2}
\end{figure*}

\begin{figure*}[t]
    \centering
    % We use \setlength to remove any horizontal space before the content.
    \setlength{\tabcolsep}{0pt}
    \begin{tabular}{m{0.04\textwidth} m{0.95\textwidth}}
        & 
        % Each \makebox is 1/7th of the line width, centering the timestamp over its frame.
        % The '%' at the end of each line is crucial to prevent unwanted spaces between the boxes.
        \makebox[\dimexpr\linewidth/5\relax][c]{0s}%
        \makebox[\dimexpr\linewidth/5\relax][c]{15s}%
        \makebox[\dimexpr\linewidth/5\relax][c]{30s}%
        \makebox[\dimexpr\linewidth/5\relax][c]{45s}%
        \makebox[\dimexpr\linewidth/5\relax][c]{60s}%
        \\[0.3em]

         &
        \includegraphics[width=\linewidth,page=1]{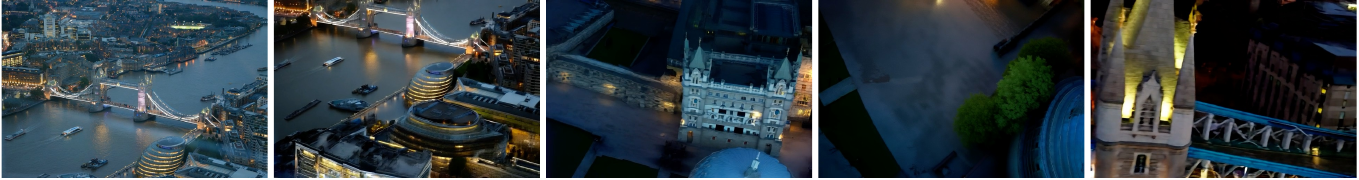} \\[0.1em]

         &
        \includegraphics[width=\linewidth,page=1]{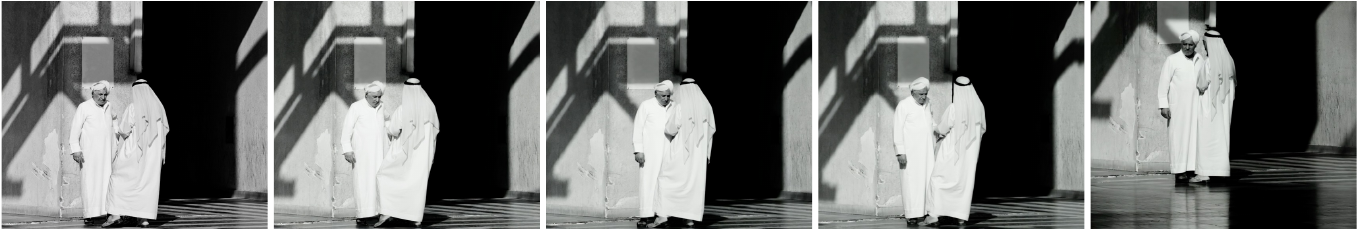} \\[0.1em]

         &
        \includegraphics[width=\linewidth,page=1]{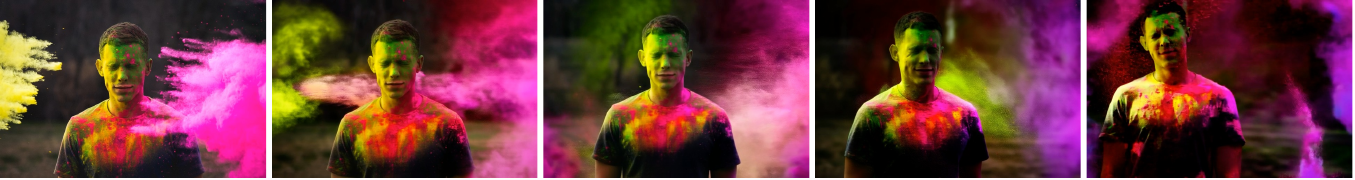} \\[0.1em]

    \end{tabular}
    \caption{Visualization of 60s videos. }

    \label{pic:app-sem-3}
    \vspace{-5ex}
\end{figure*}

We provide additional visualizations of consistency in Fig.~\ref{pic:app-sem-2}. The results show that our model maintains long-term coherence across extended sequences and successfully recovers spatial layouts that are often disrupted by complex camera motion, such as panning and zooming. Furthermore, we present additional minute-level visualizations in Fig.~\ref{pic:app-sem-3}, which highlight the model’s ability to preserve scene structure, maintain subject integrity, and avoid temporal drift over extended durations.

\label{app-section3}

\subsection{The Use of Large Language Models (LLMs)}

This thesis employs large language models (LLMs) to polish the writing and correct grammatical errors.

\end{document}